# A PDE-based log-agnostic illumination correction algorithm

Uche A. Nnolim[1]

*Department of Electronic Engineering, University of Nigeria, Nsukka, Enugu, Nigeria*

**Abstract**
*This report presents the results of a partial differential equation (PDE)-based image enhancement algorithm, for dynamic range compression and illumination correction in the absence of the logarithmic function. The proposed algorithm combines forward and reverse flows in a PDE-based formulation. The experimental results are compared with algorithms from the literature and indicate comparable performance in most cases..*

*Keywords:* Image edge contrast enhancement; illumination-reflectance model; filter kernel-based enhancement; log-agnostic dynamic range compression; power law-based illumination correction; partial differential equations

## 1. Introduction

Image enhancement is a vital preprocessing stage in image processing as it enables improved depiction of features for better analysis and extraction [1]. The feature to be enhanced include contrast, edge or colour attributes of images and several algorithms for achieving this are found in the literature and range from simple to complex algorithms [1] [2]. In the case of this work, we focus on tone mapping operation (TMO)-based algorithms, which are used to render high dynamic range (HDR) images on low dynamic range (LDR) display and print devices [1].

Examples of these TMOs include the Homomorphic [3] and Retinex [4] filters. Other algorithms include the fusion-based approaches [5], entropy-based wavelet coefficient scaling (ESWD) [6], spatially guided filtering [7], Gamma Correction (GGC) [8], Generalized Unsharp Masking (GUM) [9] algorithms [8] [10] and naturalness restoration [11]. Others include the relatively recent partial differential equations (PDE)-based enhancement by Sapiro and Caselles [12], Provenzi, et al [13], variational Retinex framework by Kimmel, et al [14], and others by [15] [16] [17] [18] [19], PDE-based Homomorphic filter (PDE-HF) [20], Dynamic Stochastic Resonance (DSR) [21], Bayesian [22] probabilistic schemes [23] and recent works such as entropy-guided Retinex-boosted AD [24], global sparse gradient guided variational Retinex (GSG-VR) [25] and variational low-light image enhancement using optimal transmission map [26].

PDE-based approaches are more versatile and flexible in addition to being able to yield intermediate and gradual results [27]. However, these algorithms do not have a means of obtaining the optimum result or an image measure-based guided approach to determine stopping time of the various PDE-based algorithms, which must be tuned to obtain good results for different images. Additionally, sub-par results are observed for several of the approaches, where there is minimal illumination normalization or considerable colour fading. Conversely, others over-enhance the images, leading to sharp divisions between light and dark regions in terms of over-exposure and under-exposure, respectively. Furthermore, most of these algorithms do not have easily implementable structure that can be developed in digital hardware

## 2. Relevant prior work

Several algorithms have attempted to solve the illumination estimation and reflectance extraction problem in a fast and compact but near accurate way and popular examples of such earlier methods are the HF and Retinex methods. The initial approaches employed the logarithmic image processing (LIP) model [1], symmetric LIP (SLIP) [8], [28] and parametric log models [29]. Additionally, variational approaches [13] [14] [15] [16] [17] [18] [19] combined with Bayesian probabilistic framework [23] have also been proposed in addition to recent works [24], [25] and [26]. Each of these methods have the strengths and weaknesses and some may exhibit high computational complexity. Additionally, these methods are not easily amenable to hardware implementation due to their structure.

The proposed tonal mapping algorithm is adaptive and guided by image metrics to ensure optimum results with the ability to process both dark and faded images while using small spatial filter kernel-based operators and avoiding logarithmic operations in a PDE-based formulation. Additionally, it attempts to solve the problem of difficulty of hardware realization by utilizing spatial filter structures to approximate these PDE processes. Furthermore, its avoidance of logarithm removes the need to implement logarithm functions for embedded mobile devices, which do not support such functions. Moreover, the algorithm has a dual-use function as it can be used to process either dark or faded, low-contrast images by adjustment of the contributions of these processes. The proposed approach also expands the discussion to generalize the algorithm to any domain that possesses contra-opposing processes for high- or low-frequency components.

3. **Proposed algorithm**
The proposed scheme can be realized in the frequency or spatial domain [30] as shown in (1);

$$I_e(x,y) = \varphi\{I_{HPF}(x,y)\} + \gamma\{I_{LPF}(x,y)\} \quad (1)$$

Where $\varphi\{\ \}$ and $\gamma\{\ \}$ are amplification and attenuation functions for the high-pass, $I_{HPF}(x,y)$ and low-pass, $I_{LPF}(x,y)$ images respectively [30] and $I_e(x,y)$ is the enhanced image. In previous work, we used the following expression to obtain the enhanced output image;

$$I_e(x,y) = I_{HPF}(x,y) + \sqrt{I_{LPF}(x,y)} \quad (2)$$

The PDE-based formulation can then be realized after derivation as;

$$\frac{\partial I(x,y,t)}{\partial t} = \lambda\left(-\nabla^2 I(x,y,t) + [D-1]^{1-k}\{I(x,y,t) + \nabla^2 I(x,y,t)\}^k - I(x,y,t)\right) + \frac{\beta(I(x,y,t)-\mu)}{\sigma} \quad (3)$$

From which we minimize the energy functional;

$$E(I) = \int_\Omega \lambda\left(-\nabla^2 I(x,y,t) + [D-1]^{1-k}\{I(x,y,t) + \nabla^2 I(x,y,t)\}^k - I(x,y,t)\right) + \frac{\beta(I(x,y,t)-\mu)}{\sigma} dx\,dy \quad (4)$$

It should be noted that the system can be generalized to any domain but for run-time purposes, we choose to utilize spatial filter or frequency domain filters. For hardware implementation, it would make sense to utilize spatial filter kernel structures, since these architectures can be easily established using hardware description language (HDL) in digital hardware such as FPGAs [30].

Experiments performed to analyze the system are shown in Fig. 1 concerning the relationship between the regularization parameter, $\lambda$ and the number of required iterations. In general, the lower the value of the parameter, the more iterations needed to attain maximum entropy, which is our optimization goal in this case. Thus, in summary, the parameter influences the execution time of the algorithm.

Additionally, some images may require much more enhancement and, thus the algorithm may be slower in attaining the optimization goal (will run for more cycles), especially if little variations are made over time rather than more drastic ones. However, this is a strength in that the approach is able to improve under-exposed regions without over-enhancing the bright regions and losing details in the process, unlike closed-form algorithms.

4. **Experiments**
In this section, we tested the proposed approach against several algorithms from the literature such as global histogram equalization (GHE), adaptive HE (AHE), contrast limited AHE (CLAHE) [31], histogram specification (HS) [32], gain offset correction (GOC1, GOC2 and GOC3) [33], piecewise linear transform (PWL) [34], linear contrast stretching (CS/LCS) [35], splitting signal alpha rooting (SSAR) [36], tonal correction (TC) [37], spatial and frequency domain HF (SHF and FDHF) [1] [38], single and multiscale Retinex with colour restoration (SSR, MSR and MSRCR) [39] [4] and GUM [9] and PDE-based formulations from earlier work [40].

We tested over 195 dark images from various sources including benchmark images used by several authors from the literature and others from the internet. Additionally, run-time comparisons were performed to assess time-complexity of the available algorithms. We performed numerical comparisons using image quality metrics such as colour enhancement factor (CEF) [41], relative mean brightness (RM), relative standard deviation (RSD) [42], relative entropy (RE), relative average gradient (RAG) [43], hue deviation index (HDI) [43] and perceptual quality metric (PQM) [21]. The relative values represent a ratio of output over input images, thus values greater than unity signify improvement, while those less than unity imply degradation. The HDI ideally should be less than one for good hue preservation while PQM should be as close to 10 as possible, where 10 is the ideal value.

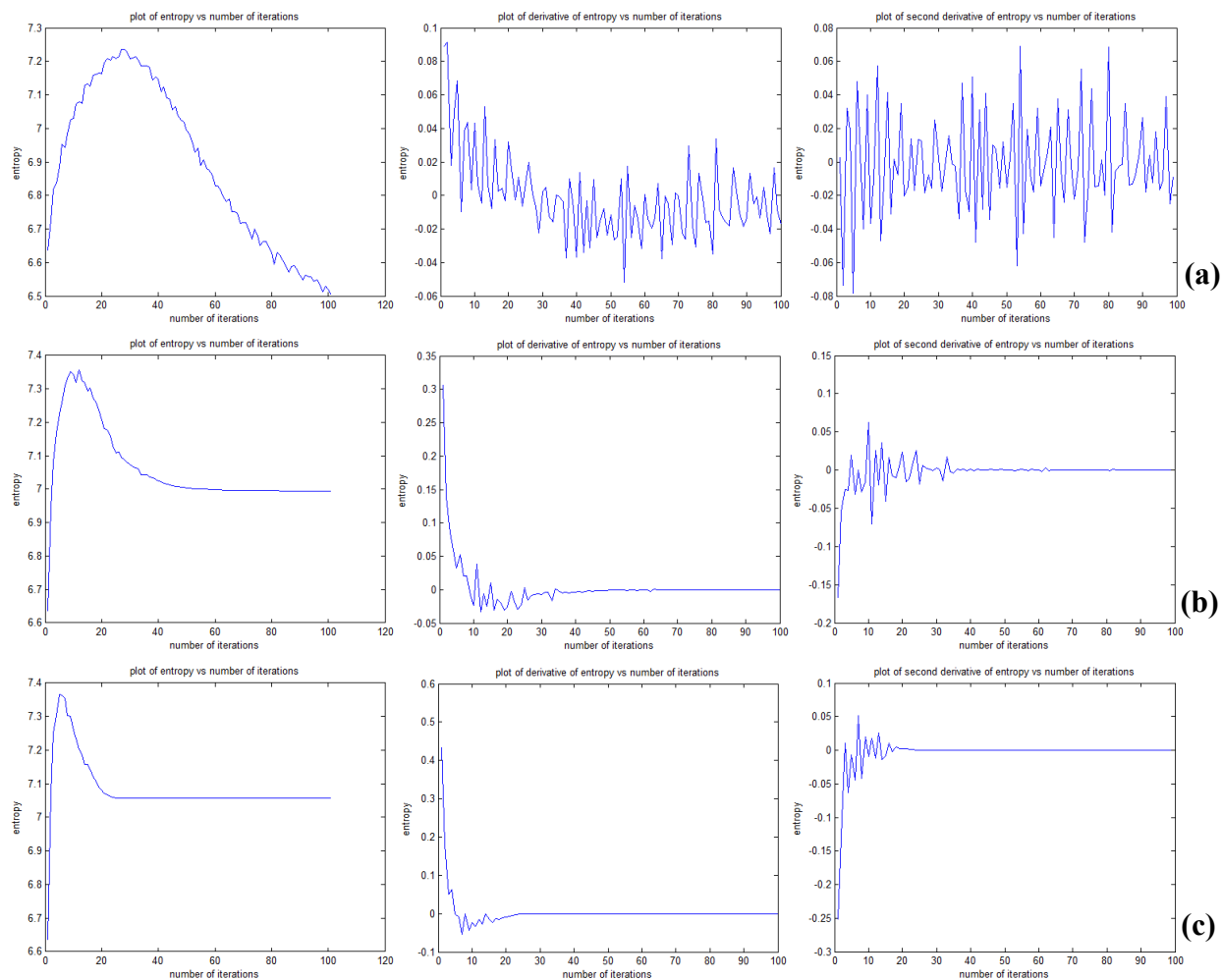

**Fig. 2** Entropy, first derivative and second derivative of entropy with respect to number of iterations for intensity channel of *Iris* image processed with PA for (a) $\lambda = 0.1$, (b) $\lambda = 0.5$ and (c) $\lambda = 1$ for 100 iterations

However, we keep in mind that numerical results based on no-reference metrics especially in the domain of illumination correction-based image enhancement is difficult to reconcile objective results with visual outcomes. This is because image colour and contrast aesthetics are not clearly quantified by numerical values, which measure structural changes but may not be precise in relating to perceptual results. We therefore focus on consistent mid-range values to represent a balanced result than obtaining extremely high or low values with usually poor visual result as will be seen in the subsequent figures.

### 4.1 Visual evaluation
In Fig.2, we provide the key to Fig. 3 where we present the visual and corresponding numerical results for the *Swan*, *Notre Dame*, *Big Ben* and *Horse* images. Based on results, the PDE-based formulations of the closed-form algorithms improve result for algorithms such as CLAHE. However, over-enhancement is observed in the swan's feathers. Most of the other algorithms darken the under-exposed regions and brighten the over-exposed regions as expected. The HF brightens the image but yields flattened and faded colours, while multi-scale Retinex algorithms give high contrast but with darkened regions and grey colouration in the white regions of the image. The worst performing algorithms are the GUM and SSR, statistical-based algorithms such as GHE, AHE, HS, which over-expose and over-enhance the entire image and distort the colours, while SSAR darkens the image. Conversely, PA brightens the dark regions of the image, while avoiding the over-exposure of bright regions, yielding a balanced result.

For the *Notre Dame* image, results are similar except that PA results in a faded sky similar to the HF, though it is less than the HF. Most of the other algorithms do not show any visible improvement or rather much worse results. Only the PDE-based CLAHE variants yield richer colours though with over-enhanced sky regions with halos and slight under-exposed dark regions while others have distorted colours. The GUM, SSR and global histogram stretching algorithms yield the worst results.

For the *Big Ben* image, the PA and HF yield similar results with faded colours in the sky regions, while the Retinex, CLAHE and others darken the image without resolution of dark regions. As usual the other statistics-based algorithms exhibit a pseudo thresholding effect by darkening under-exposed regions while brightening light areas, losing details. SSR and GUM yield over-bright images while SSAR darkens the image.

For the *Horse* image, once more, the PA and HF yield similar and best results, while the Retinex, CLAHE and others darken the image and statistics-based algorithms cannot resolve the under-exposed regions losing details in bright regions. The SSR and GUM yield over-bright images (though to a lesser extent compared with their previous performance) while SSAR once more darkens the image.

In Fig. 4 and 6 we compare visual results of PA with discrete cosine transform (DCT)-based enhancement (CES) [44], direct image contrast enhancement (DICE) [45], variational perceptually-inspired colour enhancement (VPCE) [13], image Gaussian illumination enhancement (IEGP) [46], MSR [4] and ESWD [6] from the literature [6]. Results indicate that PA surpasses several of the algorithms while being comparable to the best versions. In Fig. 5, we present additional results for PA for a wide array of images with varying levels of uneven illumination and we can see that it has problems with local contrast performance, which is expected in some cases. In Fig. 7 and 8, we also compare the PA against SLIP- based generalized gamma correction (SLIP-GGC) [8], LCS, CLAHE, local color correction (LCC) [47] [48], adaptive gamma correction (AGC) [49] and parametric log-ratio model (PLR) [50] using the figures from [8] amended with result from PA. Numerical results in Table 1 to 4 (Table 4 is from [8] amended with PA) indicate that PA yields balanced results, which are not at the extremes of too low or too high and this is in line with its visual results being subtle in enhancement unlike the other algorithms with extreme visual outcomes indicating distortions. This does not mean that PA cannot be modified to yield much higher numerical values but this would also defeat the purpose of gradual enhancement. Additionally, it should be stated that no algorithm can yield the best results for all possible images due to the uniqueness of images and their structural and perceptual features in addition to human subjective evaluation.

| Original Image | PDE_HS | PDE_GOC2 | PDE_GOC3 | PDE_PWL | PDE_GHE |
|---|---|---|---|---|---|
| PDE_CE | PDE_CS | PDE_MINMAX | PDE_AHE | PDE_CLAHE | CP_PDE_CLAHE |
| ADE1 | ADE3 | MCECR | MCECR_HF | MCECR_CLAHE | CLAHE |
| AHE | SHF | FDHF | MSR | GHE | HS |
| PWL | CS | GOC1 | GOC2 | GOC3 | RGB-IV-PA |
| HSI-PA | GUM | SSR | SSAR | TC | MSRCR |

**Fig. 2** Key to figure 3

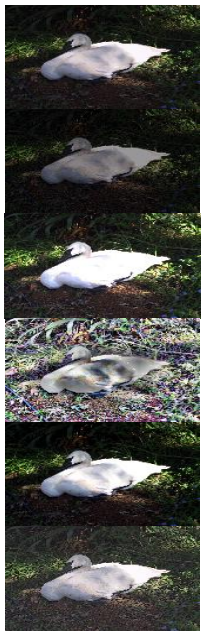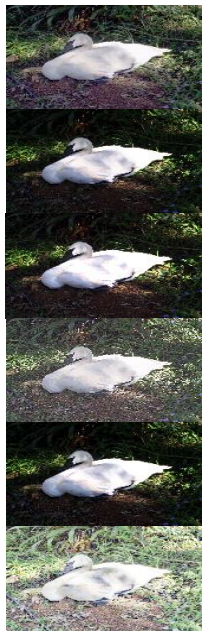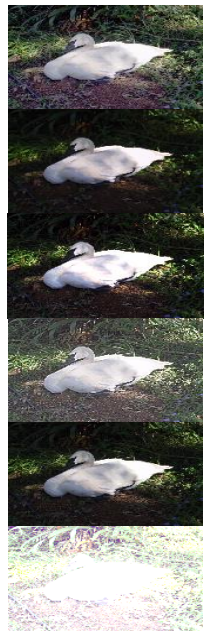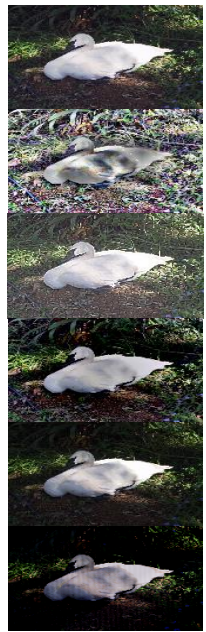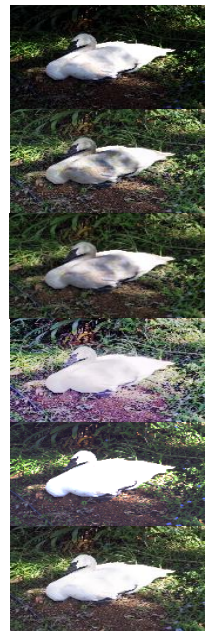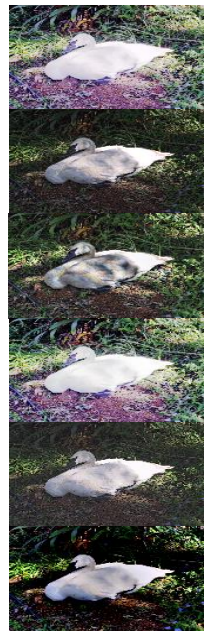

(a)

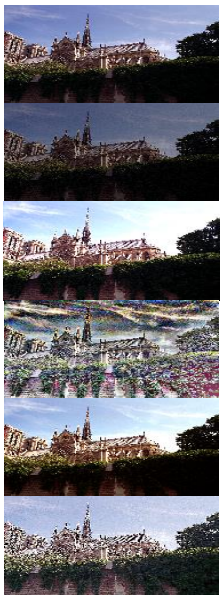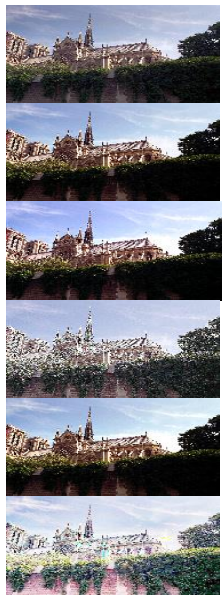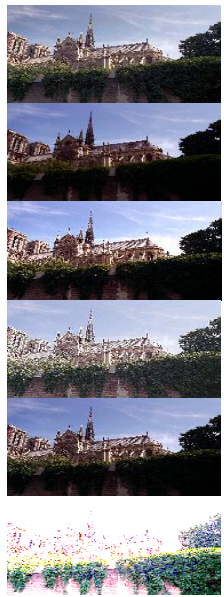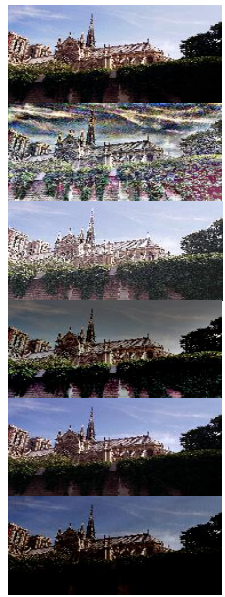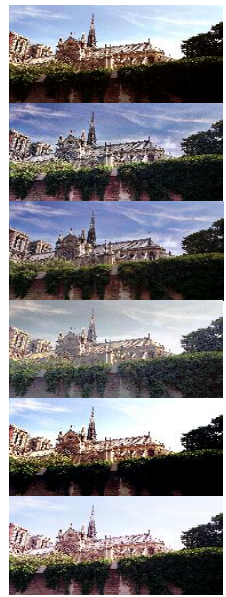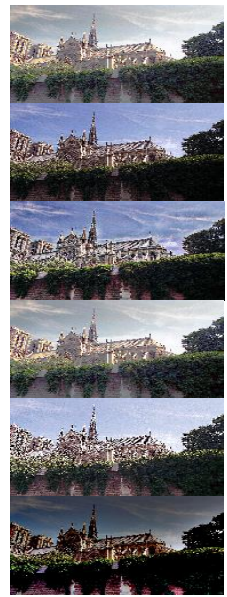

(b)

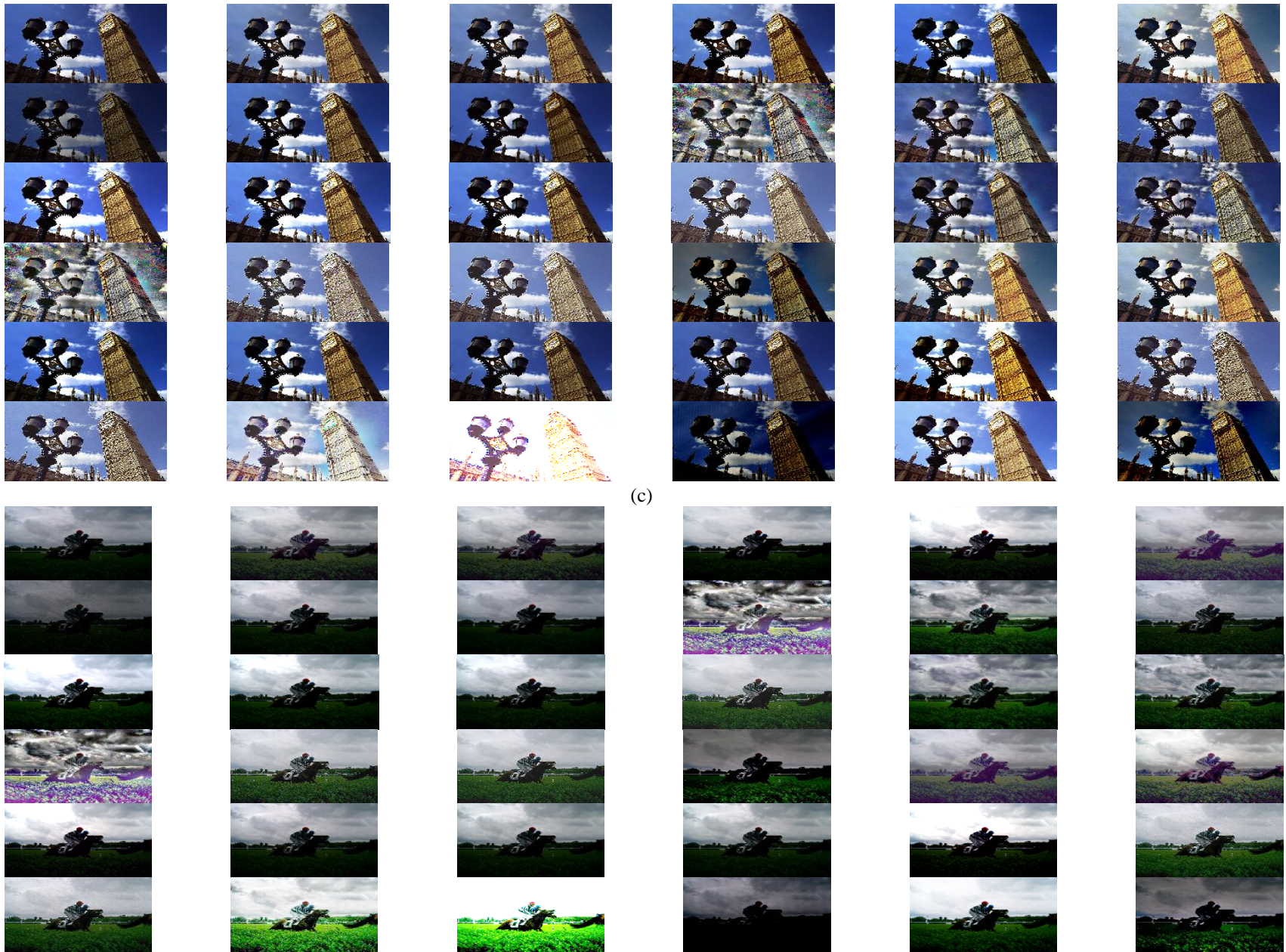

**Fig. 3** (a) Swan, (b) Notre Dame (c) Big Ben (d) H*orse* images processed with various algorithms

## 4.2 Objective evaluation

In Table 1, we present the initial obtained metrics for the sample images under test and their corresponding numerical results obtained from processing with various algorithms compared with the proposed approach (PA) in Table 2.

Table 1 Initial numerical values for the sample test images

| Swan | Notre Dame | Big Ben | Horse |
|---|---|---|---|
| Colourfulness = 15.63302 | Colourfulness = 35.278 | Colourfulness = 49.83381 | Colourfulness = 11.99735 |
| Sigma = 62.46685 | Sigma = 79.34511 | Sigma = 64.83871 | Sigma = 73.42452 |
| Mu = 52.18342 | Mu = 87.01462 | Mu = 81.48279 | Mu = 76.9868 |
| Entropy = 6.564669 | Entropy = 7.349838 | Entropy = 7.624601 | Entropy = 6.457309 |
| AG = 5.813808 | AG = 8.174997 | AG = 7.025566 | AG = 1.717612 |
| EMEC = 13.43151 | | EMEC = 94.12284 | EMEC = 94.12284 |

**Table 2** Quantitative comparison of PA with popular algorithms for (a) *Swan* (b) *Notre Dame* (c) *Big Ben* and (d) *Horse* images

(a)

| Algos\Measures | CMB | mu_o | sigma_o | ent_o | AG_o |
|---|---|---|---|---|---|
| PDE_HS | 21.41041 | 85.11685 | 63.63443 | 7.612304 | 11.00838 |
| PDE_GOC2 | 21.41041 | 85.11685 | 63.63443 | 7.612304 | 11.00838 |
| PDE_GOC3 | 19.15079 | 75.26498 | 66.61575 | 6.792743 | 7.594429 |
| PDE_PWL | 19.68157 | 52.91116 | 74.50446 | 6.479009 | 7.082669 |
| PDE_GHE | 36.67305 | 128.5168 | 71.73287 | 7.068746 | 17.83355 |
| PDE_CE | 10.38371 | 31.89485 | 38.62476 | 6.173844 | 5.474874 |
| PDE_CS | 18.87627 | 53.33921 | 72.25838 | 6.531533 | 6.860005 |
| PDE_MINMAX | 14.04993 | 47.5424 | 61.91476 | 6.292792 | 3.242516 |
| PDE_AHE | 38.53974 | 123.6227 | 68.53655 | 7.969447 | 25.05661 |
| PDE_CLAHE | 26.11403 | 79.63 | 60.62871 | 7.489354 | 12.39316 |
| CP_PDE_CLAHE | 17.46621 | 65.01815 | 60.80349 | 7.069814 | 9.788868 |
| ADE1 | 24.18329 | 75.18794 | **78.95176** | 6.169831 | 8.555037 |
| ADE3 | 21.45112 | 62.59427 | 77.6474 | 6.285439 | 7.763636 |
| MCECR | 20.35833 | 60.66397 | 75.73802 | 6.463007 | 7.394342 |
| MCECR_HF | 20.67793 | 108.9197 | 60.66652 | 7.287016 | 13.55593 |
| MCECR_CLAHE | 21.19553 | 70.48403 | 59.01643 | 7.258112 | 7.004399 |
| CLAHE | 25.45356 | 80.00689 | 59.58507 | 7.497631 | 13.24651 |
| AHE | **38.53974** | 123.6227 | 68.53655 | 7.969447 | **25.05661** |
| SHF | 21.32438 | 106.7087 | 62.67002 | 7.42941 | 17.00257 |
| FDHF | 20.42786 | 107.5572 | 59.97325 | 7.323166 | 13.52661 |
| MSR | 25.77048 | 58.80421 | 65.43866 | 6.308283 | 12.38959 |
| GHE | 36.67305 | **128.5168** | 71.73287 | 7.068746 | 17.83355 |
| HS | 38.0407 | 127.4585 | 73.90013 | **8** | 17.79277 |
| PWL | 19.68157 | 52.91116 | 74.50446 | 6.479009 | 7.082669 |
| CS | 19.97168 | 53.39783 | 75.27249 | 6.373934 | 7.168449 |
| GOC1 | 15.9899 | 52.74422 | 63.1567 | 6.568948 | 5.886273 |
| GOC2 | 15.9899 | 52.74422 | 63.1567 | 6.568948 | 5.886273 |
| GOC3 | 25.6736 | 104.934 | 74.13145 | 6.28182 | 9.922227 |
| PA_2B | 16.04195 | 76.82663 | 58.62702 | 7.037469 | 10.5224 |
| PA_1A | 16.04195 | 76.82663 | 58.62702 | 7.037469 | 10.5224 |
| GUM | 30.75108 | 163.8607 | 55.27601 | 7.67875 | 16.64929 |
| SSR | 23.46357 | 234.8064 | 31.15657 | 3.204113 | 8.721495 |
| SSAR | 14.30069 | 27.92414 | 54.50812 | 3.931229 | 3.989597 |
| TC | 25.40632 | 97.13901 | 73.3573 | 6.218762 | 10.63997 |
| MSRCR | 36.28314 | 51.58774 | 73.59523 | 5.054237 | 11.4534 |

(b)

| Algos\Measures | CMB | mu_o | sigma_o | ent_o | AG_o |
|---|---|---|---|---|---|
| PDE_HS | 24.46659 | 104.7324 | 75.92855 | 7.813199 | 8.967406 |
| PDE_GOC2 | 24.46659 | 104.7324 | 75.92855 | 7.813199 | 8.967406 |
| PDE_GOC3 | 30.30046 | 101.6759 | 92.6909 | 7.107077 | 8.974489 |
| PDE_PWL | 33.52995 | 110.5344 | 101.3294 | 6.720015 | 9.666191 |
| PDE_GHE | 22.48975 | 129.0636 | 72.4258 | 7.528004 | 10.11233 |
| PDE_CE | 21.61371 | 53.38676 | 48.52932 | 6.636064 | 6.54662 |
| PDE_CS | 31.11723 | 103.5883 | 92.74777 | 7.307866 | 9.295889 |
| PDE_MINMAX | 35.31877 | 78.78375 | 77.61752 | 7.101104 | 3.724257 |
| PDE_AHE | 37.20773 | 123.6673 | 66.02383 | 7.950476 | 25.20219 |
| PDE_CLAHE | 34.20047 | 101.0546 | 72.42216 | 7.506185 | 12.99812 |
| CP_PDE_CLAHE | 34.16708 | 95.65669 | 77.36181 | 7.580594 | 11.22214 |
| ADE1 | 30.79586 | 124.7286 | 106.5112 | 6.023334 | 10.41408 |
| ADE3 | 37.26559 | 111.7692 | 99.21694 | 6.929896 | 9.709866 |
| MCECR | 37.5011 | 107.3925 | 96.07697 | 7.158356 | 9.473395 |
| MCECR_HF | 24.29221 | 135.8742 | 80.3449 | 7.657615 | 15.08135 |
| MCECR_CLAHE | 34.54543 | 95.36644 | 72.29926 | 7.717245 | 7.682218 |
| CLAHE | 32.83629 | 100.3756 | 71.74582 | 7.500375 | 13.56241 |
| AHE | 37.20773 | 123.6673 | 66.02383 | 7.950476 | 25.20219 |
| SHF | 25.48615 | 135.4368 | 82.8546 | 7.669938 | **19.61284** |
| FDHF | 24.25264 | 134.2512 | 79.57425 | 7.665886 | 15.13771 |
| MSR | 21.05406 | 59.37361 | 53.67344 | 5.571938 | 9.142272 |
| GHE | 22.48975 | 129.0636 | 72.4258 | 7.528004 | 10.11233 |
| HS | 22.7633 | 127.5 | 73.9003 | **8** | 10.09779 |
| PWL | 33.52995 | 110.5344 | 101.3294 | 6.720015 | 9.666191 |
| CS | 31.06287 | 108.7171 | 97.06632 | 7.156449 | 9.631731 |
| GOC1 | 35.278 | 87.01462 | 79.34511 | 7.349838 | 8.174997 |
| GOC2 | 35.278 | 87.01462 | 79.34511 | 7.349838 | 8.174997 |
| GOC3 | 33.24443 | 120.4932 | **111.2518** | 5.216588 | 10.09699 |
| PA | 24.74545 | 124.2023 | 87.26492 | 7.494009 | 19.13665 |
| GUM | 33.53043 | 170.611 | 75.53924 | 7.394675 | 16.28864 |
| SSR | **46.49454** | **205.2761** | 78.58745 | 2.872876 | 13.64984 |
| SSAR | 29.29079 | 47.71003 | 56.1948 | 4.605421 | 5.308647 |
| TC | 27.52768 | 136.189 | 101.5982 | 6.528876 | 10.983 |
| MSRCR | 28.05914 | 52.1765 | 54.4629 | 5.283449 | 9.088958 |

(c)

| Algos\Measures | CMB | mu_o | sigma_o | ent_o | AG_o |
|---|---|---|---|---|---|
| PDE_HS | 41.33353 | 101.5961 | 67.18735 | 7.86657 | 7.893266 |
| PDE_GOC2 | 41.33353 | 101.5961 | 67.18735 | 7.86657 | 7.893266 |
| PDE_GOC3 | 55.87024 | 97.32826 | 76.01511 | 7.611512 | 8.237562 |
| PDE_PWL | 68.97646 | 95.15653 | 80.27077 | 7.25373 | 8.331307 |
| PDE_GHE | 45.59658 | 128.6106 | 73.16487 | 7.691652 | 9.175682 |
| PDE_CE | 30.55856 | 49.85355 | 39.86509 | 6.9508 | 5.719523 |
| PDE_CS | 55.38993 | 91.98981 | 73.59688 | 7.588292 | 7.9884 |
| PDE_MINMAX | 49.91738 | 75.97098 | 63.29997 | 7.530192 | 4.066776 |
| PDE_AHE | 41.20839 | 119.6787 | 66.66957 | 7.943151 | 18.60902 |
| PDE_CLAHE | 42.56839 | 99.47617 | 63.47974 | 7.795666 | 11.23637 |
| CP_PDE_CLAHE | 48.78202 | 89.99641 | 63.53858 | 7.714931 | 9.738881 |
| ADE1 | **76.63968** | 121.266 | 86.91034 | 6.969643 | 9.351299 |
| ADE3 | 64.82332 | 108.5387 | 81.34196 | 7.412809 | 8.864436 |
| MCECR | 61.65622 | 103.6403 | 78.84124 | 7.49019 | 8.539138 |
| MCECR_HF | 41.17095 | 139.1162 | 64.77145 | 7.726807 | 12.3757 |
| MCECR_CLAHE | 42.17686 | 94.71956 | 60.08943 | 7.758456 | 7.365915 |
| CLAHE | 43.67208 | 96.68514 | 61.74161 | 7.763405 | 11.91487 |
| AHE | 41.20839 | 119.6787 | 66.66957 | 7.943151 | **18.60902** |
| SHF | 41.75646 | 137.7719 | 66.60855 | 7.733826 | 15.42066 |
| FDHF | 40.7758 | 137.0895 | 64.11581 | 7.737171 | 12.40638 |
| MSR | 36.25065 | 69.71917 | 55.87098 | 6.510649 | 7.541222 |
| GHE | 45.59658 | 128.6106 | 73.16487 | 7.691652 | 9.175682 |
| HS | 45.91335 | 127.4779 | 73.89831 | **8** | 9.076885 |
| PWL | 68.97646 | 95.15653 | 80.27077 | 7.25373 | 8.331307 |
| CS | 57.18495 | 95.04755 | 76.28879 | 7.402076 | 8.277622 |
| GOC1 | 49.83381 | 81.48279 | 64.83871 | 7.624601 | 7.025566 |
| GOC2 | 49.83381 | 81.48279 | 64.83871 | 7.624601 | 7.025566 |
| GOC3 | 71.02618 | 117.7116 | **92.82944** | 6.245817 | 9.881299 |
| PA_2B | 42.26727 | 129.6407 | 70.63227 | 7.682708 | 16.59522 |
| PA_1A | 42.26727 | 129.6407 | 70.63227 | 7.682708 | 16.59522 |
| GUM | 42.19889 | 179.4507 | 59.37748 | 7.574447 | 11.675 |
| SSR | 51.7097 | **233.9467** | 51.76643 | 1.909429 | 7.851679 |
| SSAR | 39.59303 | 37.65087 | 48.77306 | 5.352086 | 5.184091 |
| TC | 58.78623 | 144.173 | 80.80218 | 7.093402 | 9.588547 |
| MSRCR | 55.10485 | 61.47176 | 63.31255 | 6.144844 | 7.80667 |

(d)

| Algos\Measures | CMB | mu_o | sigma_o | ent_o | AG_o |
|---|---|---|---|---|---|
| PDE_HS | 14.39133 | 99.08217 | 72.66473 | 7.611935 | 3.746073 |
| PDE_GOC2 | 14.39133 | 99.08217 | 72.66473 | 7.611935 | 3.746073 |
| PDE_GOC3 | 11.29153 | 95.65477 | 90.88007 | 6.340069 | 1.970514 |
| PDE_PWL | 11.65842 | 112.4371 | 104.6785 | 6.198049 | 2.326264 |
| PDE_GHE | 22.01073 | 116.1428 | 58.05779 | 6.760153 | 4.086882 |
| PDE_CE | 7.884665 | 47.40972 | 44.73713 | 5.911963 | 2.386313 |
| PDE_CS | 12.10946 | 85.81693 | 81.738 | 6.542314 | 1.916801 |
| PDE_MINMAX | 11.89701 | 76.00692 | 73.561 | 6.164958 | 1.204559 |
| PDE_AHE | 47.13342 | 133.6771 | 54.90483 | 7.76614 | 11.51669 |
| PDE_CLAHE | 26.78164 | 84.53481 | 70.30413 | 6.876912 | 4.020784 |
| CP_PDE_CLAHE | 15.21239 | 83.25001 | 73.40733 | 6.642108 | 3.147387 |
| ADE1 | 19.12352 | 117.4214 | 109.3717 | 5.672184 | 2.412008 |
| ADE3 | 14.67429 | 102.3854 | 97.06013 | 6.351584 | 2.225564 |
| MCECR | 13.91154 | 97.65126 | 92.94595 | 6.583243 | 2.130172 |
| MCECR_HF | 23.39278 | 122.1327 | 79.73324 | 7.111243 | 4.970293 |
| MCECR_CLAHE | 25.37082 | 81.77472 | 71.20501 | 6.978516 | 2.555353 |
| CLAHE | 23.73371 | 83.95083 | 68.66829 | 6.764479 | 4.265469 |
| AHE | 47.13342 | 133.6771 | 54.90483 | 7.76614 | **11.51669** |
| SHF | 25.82467 | 118.319 | 80.81049 | 7.330449 | 7.40167 |
| FDHF | 23.10006 | 120.4601 | 78.6659 | 7.118756 | 5.06522 |
| MSR | 30.61708 | 59.19881 | 51.49067 | 5.266035 | 3.213714 |
| GHE | 22.01073 | 116.1428 | 58.05779 | 6.760153 | 4.086882 |
| HS | 27.14742 | 127.4739 | 73.89467 | **8** | 6.390259 |
| PWL | 11.65842 | 112.4371 | 104.6785 | 6.198049 | 2.326264 |
| CS | 12.20694 | 88.2591 | 84.09863 | 6.551213 | 1.967786 |
| GOC1 | 12.15898 | 88.68082 | 84.42666 | 6.514779 | 1.977533 |
| GOC2 | 12.15898 | 88.68082 | 84.42666 | 6.514779 | 1.977533 |
| GOC3 | 12.011 | 119.6163 | **114.1335** | 4.890023 | 2.303063 |
| PA_2B | 26.93677 | 112.6386 | 85.44763 | 7.230468 | 5.275181 |
| PA_1A | 26.93677 | 112.6386 | 85.44763 | 7.230468 | 5.275181 |
| GUM | 72.85171 | 149.4588 | 92.856 | 6.501869 | 10.1732 |
| SSR | **104.8461** | **173.5025** | 104.4962 | 2.709715 | 9.236858 |
| SSAR | 8.102315 | 47.36964 | 51.38798 | 4.524359 | 1.110965 |
| TC | 25.64745 | 125.3163 | 108.3916 | 5.924862 | 2.977004 |
| MSRCR | 33.68869 | 53.31446 | 49.97375 | 5.038495 | 2.968169 |

**Table 3** Quantitative (relative) comparison of PA with popular algorithms for (a) *Swan* (b) *Notre Dame* (c) *Big Ben* (d) *Horse* images

(a)

| Algos\Measures | RC | F | PQM | REMEC | RM | RSD | RE | RAG | HDI | EMEC_2 |
|---|---|---|---|---|---|---|---|---|---|---|
| PDE_HS | 1.369563 | 0.636212 | 9.215336 | 1.247953 | 1.631109 | 1.018691 | 1.159587 | 1.893488 | 6.989006 | 16.76189 |
| PDE_GOC2 | 1.369563 | 0.636212 | 9.215336 | 1.247953 | 1.631109 | 1.018691 | 1.159587 | 1.893488 | 6.989006 | 16.76189 |
| PDE_GOC3 | 1.225021 | 0.788486 | 9.904711 | 0.698589 | 1.442316 | 1.066418 | 1.034743 | 1.306275 | 7.320744 | 9.383101 |
| PDE_PWL | 1.258974 | 1.402977 | 10.00693 | 10.29938 | 1.013946 | 1.192704 | 0.986951 | 1.21825 | 5.112686 | 138.3362 |
| PDE_GHE | 2.345871 | 0.535439 | 8.085046 | 2.073571 | 2.46279 | 1.148335 | 1.076786 | 3.067447 | 11.31385 | 27.85118 |
| PDE_CE | 0.664216 | 0.625525 | 10.70477 | **13.08342** | 0.611207 | 0.618324 | 0.940466 | 0.941702 | 7.056079 | 175.7301 |
| PDE_CS | 1.207461 | 1.309071 | 10.09095 | 1.700594 | 1.022149 | 1.156748 | 0.994952 | 1.17995 | 3.841516 | 22.84154 |
| PDE_MINMAX | 0.898734 | 1.078303 | 9.667047 | 0.794077 | 0.911063 | 0.991162 | 0.958585 | 0.557727 | 4.609464 | 10.66566 |
| PDE_AHE | 2.465277 | 0.508135 | 7.568375 | 3.831895 | 2.369004 | 1.097167 | 1.213991 | 4.309845 | 14.60396 | 51.46812 |
| PDE_CLAHE | 1.67044 | 0.617324 | 8.944542 | 1.205058 | 1.525963 | 0.970574 | 1.140858 | 2.131676 | 6.402671 | 16.18574 |
| CP_PDE_CLAHE | 1.117264 | 0.760424 | 9.643628 | 1.122367 | 1.245954 | 0.973372 | 1.076949 | 1.683727 | 5.286975 | 15.07507 |
| ADE1 | 1.546936 | 1.108687 | 9.578401 | 0.994625 | 1.440839 | **1.263899** | 0.939854 | 1.471503 | 2.18659 | 13.35931 |
| ADE3 | 1.372167 | 1.288109 | 9.81832 | 4.734788 | 1.199505 | 1.243018 | 0.957465 | 1.335379 | 3.320813 | 63.59534 |
| MCECR | 1.302264 | 1.264534 | 9.952609 | 3.954112 | 1.162514 | 1.212451 | 0.984514 | 1.271859 | 2.993008 | 53.10968 |
| MCECR_HF | 1.322708 | 0.451882 | 8.866753 | 0.693553 | 2.087247 | 0.971179 | 1.110036 | 2.331679 | 5.240937 | 9.315461 |
| MCECR_CLAHE | 1.355817 | 0.660828 | 9.53471 | 0.838467 | 1.350698 | 0.944764 | 1.105633 | 1.204787 | 7.422742 | 11.26187 |
| CLAHE | 1.628192 | 0.593445 | 8.840716 | 1.225813 | 1.533186 | 0.953867 | 1.142119 | 2.278457 | 6.69303 | 16.46451 |
| AHE | **2.465277** | 0.508135 | 7.568375 | 3.831895 | 2.369004 | 1.097167 | 1.213991 | **4.309845** | 14.60396 | 51.46812 |
| SHF | 1.36406 | 0.492213 | 8.463246 | 0.946139 | 2.044878 | 1.003252 | 1.131727 | 2.924515 | 7.645274 | 12.70807 |
| FDHF | 1.306712 | 0.447208 | 8.913402 | 0.701319 | 2.061137 | 0.960081 | 1.115542 | 2.326636 | 5.14373 | 9.419773 |
| MSR | 1.648464 | 0.973854 | 8.543244 | | 1.126875 | 1.047574 | 0.960945 | 2.131063 | 9.087314 | |
| GHE | 2.345871 | 0.535439 | 8.085046 | 2.073571 | **2.46279** | 1.148335 | 1.076786 | 3.067447 | 11.31385 | **27.85118** |
| HS | 2.433355 | 0.573 | 8.217422 | 3.256731 | 2.44251 | 1.18303 | **1.218645** | 3.060433 | 10.83301 | 43.7428 |
| PWL | 1.258974 | 1.402977 | **10.00693** | 10.29938 | 1.013946 | 1.192704 | 0.986951 | 1.21825 | 5.112686 | 138.3362 |
| CS | 1.277531 | 1.419 | 9.959905 | | 1.023272 | 1.204999 | 0.970945 | 1.233004 | 5.314325 | |
| GOC1 | 1.022828 | 1.01134 | 10.46848 | 1.006673 | 1.010747 | 1.011043 | 1.000652 | 1.012464 | 2.870382 | 13.52114 |
| GOC2 | 1.022828 | 1.01134 | 10.46848 | 1.006673 | 1.010747 | 1.011043 | 1.000652 | 1.012464 | 2.870382 | 13.52114 |
| GOC3 | 1.642267 | 0.700361 | 9.175185 | 0.612586 | 2.010868 | 1.186733 | 0.956914 | 1.706666 | 13.69751 | 8.227953 |
| PA | 1.026158 | 0.598298 | 9.441058 | 0.912874 | 1.472242 | 0.93853 | 1.072022 | 1.809898 | 6.314627 | 12.26128 |
| GUM | 1.967059 | 0.249363 | 8.3413 | 0.692045 | 3.140091 | 0.884885 | 1.169709 | 2.86375 | 7.819348 | 9.295204 |
| SSR | 1.500898 | 0.055287 | 8.660112 | 0.296004 | 4.499635 | 0.49877 | 0.488084 | 1.500135 | 9.328737 | 3.975786 |
| SSAR | 0.914774 | **1.422905** | 10.10053 | | 0.535115 | 0.872593 | 0.598847 | 0.686228 | 12.88947 | |
| TC | 1.62517 | 0.740843 | 9.147615 | 0.867176 | 1.861492 | 1.17434 | 0.947308 | 1.830121 | **1.139818** | 11.64748 |
| MSRCR | 2.320929 | 1.404061 | 8.303884 | | 0.988585 | 1.178148 | 0.769915 | 1.970034 | 10.10144 | |

(b)

| Algos\Measures | RC | F | PQM | RM | RSD | RE | RAG | HDI |
|---|---|---|---|---|---|---|---|---|
| PDE_HS | 0.693537 | 0.760818 | 8.700638 | 1.203619 | 0.956941 | 1.063044 | 1.096931 | 3.2868 |
| PDE_GOC2 | 0.693537 | 0.760818 | 8.700638 | 1.203619 | 0.956941 | 1.063044 | 1.096931 | 3.2868 |
| PDE_GOC3 | 0.858905 | 1.167906 | 8.755461 | 1.168492 | 1.168199 | 0.966971 | 1.097797 | 3.357777 |
| PDE_PWL | 0.950449 | 1.283884 | 8.443598 | 1.270297 | 1.277072 | 0.914308 | 1.182409 | 13.3033 |
| PDE_GHE | 0.637501 | 0.561739 | 8.079307 | 1.48324 | 0.912795 | 1.024241 | 1.236982 | 10.25374 |
| PDE_CE | 0.612668 | 0.609715 | 9.148268 | 0.613538 | 0.611623 | 0.902886 | 0.80081 | 4.454588 |
| PDE_CS | 0.882058 | 1.147753 | 8.70612 | 1.190469 | 1.168916 | 0.994289 | 1.137112 | 5.114248 |
| PDE_MINMAX | 1.001155 | 1.056902 | 8.394005 | 0.905408 | 0.978227 | 0.966158 | 0.455567 | 1.254058 |
| PDE_AHE | 1.054701 | 0.48719 | 5.921435 | 1.421224 | 0.83211 | 1.081721 | 3.082838 | 18.66341 |
| PDE_CLAHE | 0.969456 | 0.717363 | 7.87316 | 1.161352 | 0.912749 | 1.021272 | 1.589985 | 2.995098 |
| CP_PDE_CLAHE | 0.968509 | 0.864749 | 8.388061 | 1.099317 | 0.975004 | 1.031396 | 1.372739 | 1.85013 |
| ADE1 | 0.872948 | 1.257119 | 8.210378 | 1.433421 | 1.342379 | 0.819519 | 1.273894 | 6.158379 |
| ADE3 | 1.056341 | 1.217311 | 8.511933 | 1.284487 | 1.250448 | 0.942864 | 1.187752 | 2.694105 |
| MCECR | 1.063017 | 1.188 | 8.630466 | 1.234189 | 1.210875 | 0.973947 | 1.158825 | 2.023707 |
| MCECR_HF | 0.688594 | 0.656646 | 7.727534 | 1.56151 | 1.012601 | 1.041875 | 1.844814 | 3.751029 |
| MCECR_CLAHE | 0.979234 | 0.757572 | 9.096887 | 1.095982 | 0.9112 | 1.049988 | 0.939721 | 5.971683 |
| CLAHE | 0.930787 | 0.708789 | 7.818679 | 1.153548 | 0.904225 | 1.020482 | 1.659011 | 3.192205 |
| AHE | 1.054701 | 0.48719 | 5.921435 | 1.421224 | 0.83211 | 1.081721 | **3.082838** | 18.66341 |
| SHF | 0.722438 | 0.700565 | 6.972947 | 1.556483 | 1.044231 | 1.043552 | 2.399125 | 6.349478 |
| FDHF | 0.687472 | 0.651897 | 7.779067 | 1.542858 | 1.002888 | 1.043001 | 1.851708 | 3.786277 |
| MSR | 0.596804 | 0.670621 | 8.127166 | 0.682341 | 0.676456 | 0.758104 | 1.118321 | 15.19259 |
| GHE | 0.637501 | 0.561739 | 8.079307 | 1.48324 | 0.912795 | 1.024241 | 1.236982 | 10.25374 |
| HS | 0.645255 | 0.592017 | 8.316356 | 1.465271 | 0.931378 | **1.088459** | 1.235204 | 9.923013 |
| PWL | 0.950449 | 1.283884 | 8.443598 | 1.270297 | 1.277072 | 0.914308 | 1.182409 | 13.3033 |
| CS | 0.880517 | 1.197819 | 8.583178 | 1.249411 | 1.223343 | 0.973688 | 1.178194 | 7.157877 |
| GOC1 | 1 | 1 | 9.0031 | 1 | 1 | 1 | 1 | **0** |
| GOC2 | 1 | 1 | 9.0031 | 1 | 1 | 1 | 1 | **0** |
| GOC3 | 0.942356 | **1.419723** | 7.99827 | 1.384747 | **1.402126** | 0.709756 | 1.235106 | 15.94169 |
| PA_1A | 0.701441 | 0.847426 | 7.234434 | 1.427373 | 1.099815 | 1.019616 | 2.340876 | 2.566798 |
| GUM | 0.950463 | 0.462264 | 6.882727 | 1.960716 | 0.952034 | 1.0061 | 1.992495 | 5.490489 |
| SSR | **1.317947** | 0.415834 | 5.661471 | **2.359099** | 0.990451 | 0.390876 | 1.669706 | 15.93422 |
| SSAR | 0.830285 | 0.914818 | **9.221785** | 0.548299 | 0.708233 | 0.626602 | 0.649376 | 11.76553 |
| TC | 0.780307 | 1.047567 | 8.122459 | 1.565128 | 1.28046 | 0.888302 | 1.343487 | 1.350291 |
| MSRCR | 0.795372 | 0.785739 | 8.104235 | 0.599629 | 0.686405 | 0.718852 | 1.1118 | 12.62348 |

(c)

| Algos\Measures | RC | F | PQM | REMEC | RM | RSD | RE | RAG | HDI | EMEC_2 |
|---|---|---|---|---|---|---|---|---|---|---|
| PDE_HS | 0.829427 | 0.861182 | 9.127733 | 0.647361 | 1.246841 | 1.036223 | 1.031735 | 1.123506 | 3.585806 | 60.9315 |
| PDE_GOC2 | 0.829427 | 0.861182 | 9.127733 | 0.647361 | 1.246841 | 1.036223 | 1.031735 | 1.123506 | 3.585806 | 60.9315 |
| PDE_GOC3 | 1.121131 | 1.150689 | 8.981254 | 1.006669 | 1.194464 | 1.172372 | 0.998283 | 1.172512 | 2.054944 | 94.75051 |
| PDE_PWL | 1.38413 | 1.312421 | 8.859528 | | 1.167811 | 1.238007 | 0.951359 | 1.185856 | 4.985989 | |
| PDE_GHE | 0.914973 | 0.806725 | 8.602648 | 0.138368 | 1.578378 | 1.128413 | 1.008794 | 1.306042 | 9.123589 | 13.02356 |
| PDE_CE | 0.613209 | 0.617855 | **9.920378** | **1.425868** | 0.611829 | 0.614835 | 0.911628 | 0.814101 | 1.595194 | **134.2068** |
| PDE_CS | 1.111493 | 1.141238 | 9.071707 | 1.082313 | 1.128948 | 1.135076 | 0.995238 | 1.137047 | 1.775194 | 101.8703 |
| PDE_MINMAX | 1.001677 | 1.022248 | 8.449016 | 0.984505 | 0.932356 | 0.976268 | 0.987618 | 0.578854 | **0.69743** | 92.6644 |
| PDE_AHE | 0.826916 | 0.719839 | 6.763662 | 0.245432 | 1.46876 | 1.028237 | 1.041779 | 2.648758 | 18.3554 | 23.10079 |
| PDE_CLAHE | 0.854207 | 0.785142 | 8.399892 | 0.175629 | 1.220824 | 0.979041 | 1.022436 | 1.599354 | 4.277394 | 16.53068 |
| CP_PDE_CLAHE | 0.978894 | 0.869455 | 8.882895 | 0.192719 | 1.104484 | 0.979948 | 1.011847 | 1.386206 | 1.421066 | 18.13924 |
| ADE1 | **1.537905** | 1.207261 | 8.51988 | 1.004228 | 1.48824 | 1.340408 | 0.914099 | 1.331038 | 3.348789 | 94.52078 |
| ADE3 | 1.30079 | 1.181521 | 8.766637 | 1.078936 | 1.332045 | 1.254528 | 0.972223 | 1.26174 | 3.133409 | 101.5525 |
| MCECR | 1.237237 | 1.162453 | 8.8801 | 1.0773 | 1.271929 | 1.215959 | 0.982371 | 1.215438 | 2.670539 | 101.3986 |
| MCECR_HF | 0.826165 | 0.584503 | 8.302642 | 0.120801 | 1.707307 | 0.998963 | 1.013405 | 1.761523 | 3.048799 | 11.37013 |
| MCECR_CLAHE | 0.84635 | 0.738845 | 9.174231 | 0.135545 | 1.162449 | 0.926752 | 1.017556 | 1.048444 | 3.175443 | 12.7579 |
| CLAHE | 0.876354 | 0.764176 | 8.324198 | 0.178226 | 1.186571 | 0.952234 | 1.018205 | 1.69593 | 4.169494 | 16.7751 |
| AHE | 0.826916 | 0.719839 | 6.763662 | 0.245432 | 1.46876 | 1.028237 | **1.041779** | **2.648758** | 18.3554 | 23.10079 |
| SHF | 0.837914 | 0.624161 | 7.963148 | 0.296767 | 1.69081 | 1.027296 | 1.014325 | 2.194935 | 4.642079 | 27.9326 |
| FDHF | 0.818236 | 0.581197 | 8.355002 | 0.121922 | 1.682435 | 0.988851 | 1.014764 | 1.765891 | 2.953714 | 11.47565 |
| MSR | 0.727431 | 0.867796 | 8.769987 | | 0.855631 | 0.861692 | 0.8539 | 1.073397 | 9.426818 | |
| GHE | 0.914973 | 0.806725 | 8.602648 | 0.138368 | 1.578378 | 1.128413 | 1.008794 | 1.306042 | 9.123589 | 13.02356 |
| HS | 0.921329 | 0.830293 | 8.811686 | 0.490795 | 1.564476 | 1.139725 | 1.049235 | 1.291979 | 8.849393 | 46.19503 |
| PWL | 1.38413 | 1.312421 | 8.859528 | | 1.167811 | 1.238007 | 0.951359 | 1.185856 | 4.985989 | |
| CS | 1.147513 | 1.1868 | 8.923026 | | 1.166474 | 1.176593 | 0.970815 | 1.178214 | 3.612152 | |
| GOC1 | 1 | 1 | 9.380909 | 1 | 1 | 1 | 1 | 1 | **0** | 94.12284 |
| GOC2 | 1 | 1 | 9.380909 | 1 | 1 | 1 | 1 | 1 | **0** | 94.12284 |
| GOC3 | 1.425261 | **1.418892** | 8.031775 | | 1.444619 | **1.431698** | 0.819166 | 1.406477 | 8.152909 | |
| PA | 0.848165 | 0.745868 | 7.850542 | 1.012947 | 1.59102 | 1.089353 | 1.007621 | 2.362118 | 2.559008 | 95.34145 |
| GUM | 0.846792 | 0.380799 | 8.148471 | 0.291564 | 2.202314 | 0.915772 | 0.993422 | 1.661788 | 6.834022 | 27.44282 |
| SSR | 1.037643 | 0.222012 | 6.830193 | 0.94169 | **2.871119** | 0.798388 | 0.25043 | 1.117587 | 23.86199 | 88.63454 |
| SSAR | 0.794501 | 1.224565 | 9.805421 | | 0.462071 | 0.752221 | 0.70195 | 0.737889 | 6.256646 | |
| TC | 1.179645 | 0.877727 | 8.533475 | 0.985197 | 1.769367 | 1.246203 | 0.930331 | 1.364808 | 2.06565 | 92.72955 |
| MSRCR | 1.105772 | 1.263866 | 8.574126 | | 0.754414 | 0.976462 | 0.805923 | 1.11118 | 7.668157 | |

(d)

| Algos\Measures | RC | F | PQM | RM | RSD | RE | RAG | HDI |
|---|---|---|---|---|---|---|---|---|
| PDE_HS | 1.199543 | 0.761002 | 10.87293 | 1.287002 | 0.989652 | **1.178809** | 2.180977 | 7.99827 |
| PDE_GOC2 | 1.199543 | 0.761002 | 10.87293 | 1.287002 | 0.989652 | 1.178809 | 2.180977 | 7.99827 |
| PDE_GOC3 | 0.941169 | 1.233005 | 11.71787 | 1.242483 | 1.237735 | 0.981844 | 1.14724 | 1.683549 |
| PDE_PWL | 0.97175 | 1.391679 | 11.29287 | 1.460472 | 1.425661 | 0.95985 | 1.354359 | 13.96247 |
| PDE_GHE | 1.834633 | 0.414441 | 10.27025 | 1.508606 | 0.790714 | 1.046899 | 2.379397 | 16.23942 |
| PDE_CE | 0.657201 | 0.602841 | 11.95958 | 0.615816 | 0.609294 | 0.915546 | 1.38932 | 4.445607 |
| PDE_CS | 1.009345 | 1.111755 | 11.87911 | 1.114697 | 1.113225 | 1.013164 | 1.115969 | 1.076393 |
| PDE_MINMAX | 0.991636 | 1.016661 | 9.636885 | 0.987272 | 1.001859 | 0.954726 | 0.701298 | 1.271472 |
| PDE_AHE | 3.928653 | 0.322031 | 8.635156 | 1.736364 | 0.747772 | 1.20269 | 6.70506 | 24.80391 |
| PDE_CLAHE | 2.232296 | 0.834949 | 10.65871 | 1.098043 | 0.957502 | 1.064981 | 2.340915 | 4.978221 |
| CP_PDE_CLAHE | 1.267979 | 0.924334 | 11.43345 | 1.081354 | 0.999766 | 1.028619 | 1.83242 | 1.46052 |
| ADE1 | 1.593979 | 1.454779 | 11.18268 | 1.525215 | 1.489581 | 0.878413 | 1.40428 | 5.058383 |
| ADE3 | 1.223128 | 1.313947 | 11.48456 | 1.329909 | 1.321904 | 0.983627 | 1.295731 | 3.364533 |
| MCECR | 1.159551 | 1.263331 | 11.64886 | 1.268416 | 1.265871 | 1.019503 | 1.240194 | 2.013526 |
| MCECR_HF | 1.949829 | 0.743329 | 10.3957 | 1.586411 | 1.085921 | 1.10127 | 2.893722 | 2.998113 |
| MCECR_CLAHE | 2.114702 | 0.885393 | 10.46383 | 1.062191 | 0.969772 | 1.080716 | 1.487736 | 4.075951 |
| CLAHE | 1.978246 | 0.802087 | 10.57824 | 1.090457 | 0.935223 | 1.047569 | 2.483371 | 3.57745 |
| AHE | 3.928653 | 0.322031 | 8.635156 | 1.736364 | 0.747772 | 1.20269 | 6.70506 | **24.80391** |
| SHF | 2.152531 | 0.788162 | **10.06284** | 1.536873 | 1.100593 | 1.135217 | 4.309279 | 4.310034 |
| FDHF | 1.92543 | 0.733608 | 10.59391 | 1.564685 | 1.071385 | 1.102434 | 2.948989 | 3.064269 |
| MSR | 2.551987 | 0.639555 | 10.55708 | 0.768948 | 0.701274 | 0.815515 | 1.871036 | 18.27145 |
| GHE | 1.834633 | 0.414441 | 10.27025 | 1.508606 | 0.790714 | 1.046899 | 2.379397 | 16.23942 |
| HS | 2.262785 | 0.611701 | 9.796438 | 1.655789 | 1.006403 | 1.238906 | 3.720432 | 21.57824 |
| PWL | 0.97175 | 1.391679 | 11.29287 | 1.460472 | 1.425661 | 0.95985 | 1.354359 | 13.96247 |
| CS | 1.017469 | 1.144333 | 11.82769 | 1.146419 | 1.145375 | 1.014542 | 1.145652 | 1.568112 |
| GOC1 | 1.013472 | 1.147793 | 11.81415 | 1.151896 | 1.149843 | 1.0089 | 1.151327 | 2.348662 |
| GOC2 | 1.013472 | 1.147793 | 11.81415 | 1.151896 | 1.149843 | 1.0089 | 1.151327 | 2.348662 |
| GOC3 | 1.001138 | **1.555142** | 10.9191 | 1.553725 | **1.554433** | 0.757285 | 1.340852 | 17.46785 |
| PA_2B | 2.245227 | 0.92565 | 10.65496 | 1.46309 | 1.163748 | 1.119734 | 3.071229 | 3.828653 |
| PA_1A | 2.245227 | 0.92565 | 10.65496 | 1.46309 | 1.163748 | 1.119734 | 3.071229 | 3.828653 |
| GUM | 6.072317 | 0.823821 | 8.613029 | 1.941356 | 1.264646 | 1.006901 | **5.922875** | 8.354441 |
| SSR | **8.739106** | 0.898731 | 7.417722 | **2.253666** | 1.423179 | 0.419635 | 5.377731 | 21.01206 |
| SSAR | 0.675342 | 0.796081 | 12.14461 | 0.615296 | 0.699875 | 0.700657 | 0.646808 | 16.43802 |
| TC | 2.13776 | 1.338806 | 10.84438 | 1.627764 | 1.476232 | 0.917543 | 1.733222 | 0.554686 |
| MSRCR | 2.808011 | 0.668918 | 10.54921 | 0.692514 | 0.680614 | 0.780278 | 1.728078 | 14.26765 |

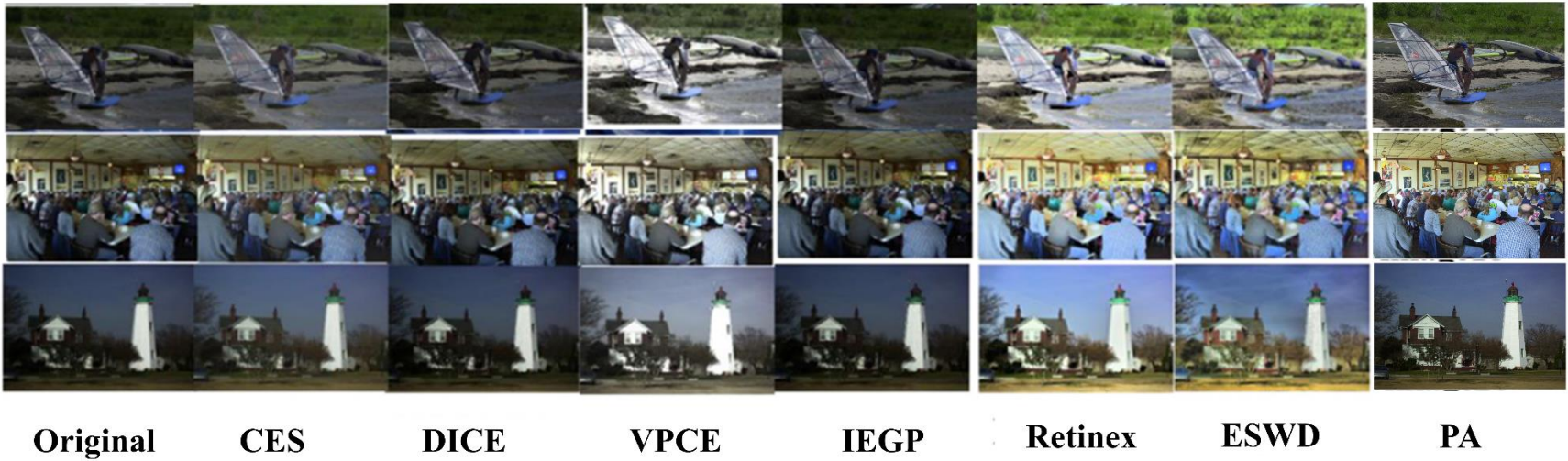

**Fig. 4** (a) Swan, (b) Notre Dame (c) Big Ben (d) H*orse* images processed with various algorithms and (e) key to figures

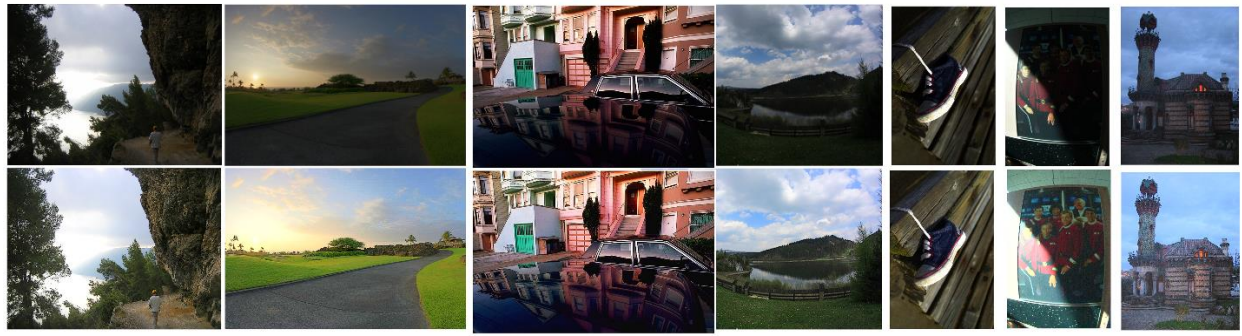

**Fig. 5** Additional results using PA

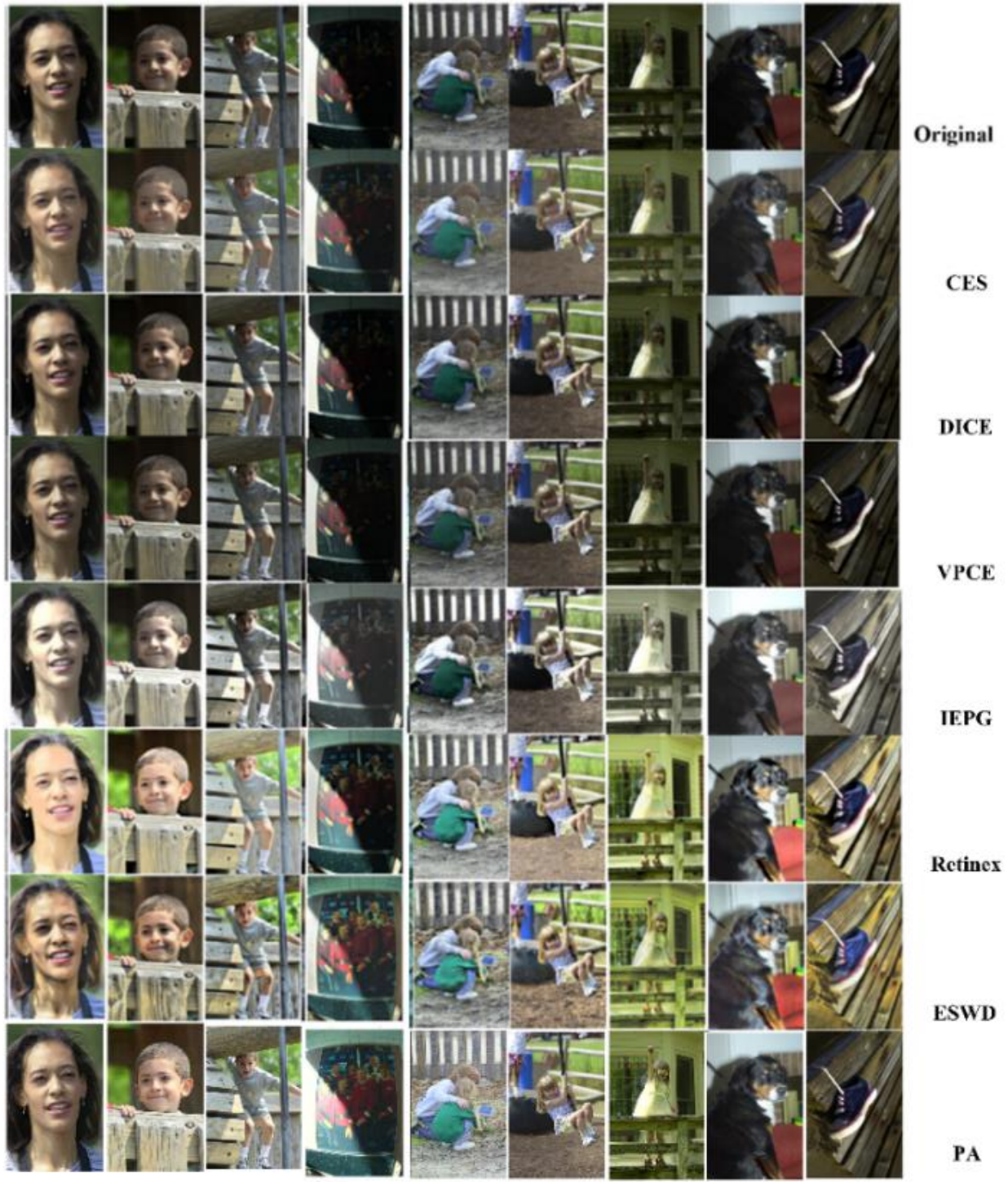

**Fig. 6** Figure from [6] amended with the results from PA (last row) for visual comparison

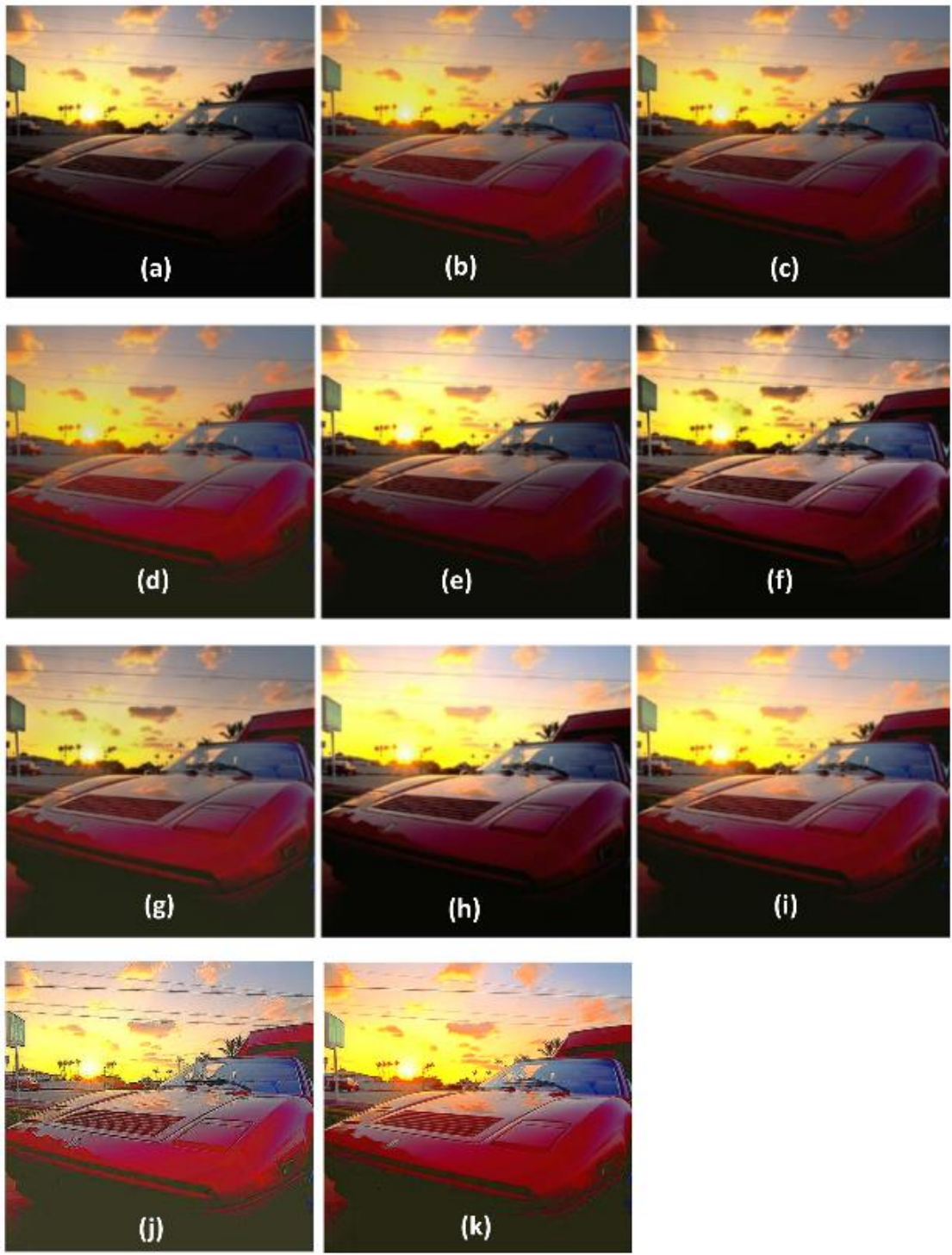

**Fig. 7** Amended figure from [8] of Ferrari image showing (a) original image (b) to (d) SLIP using various values (e) LCS (f) CLAHE (g) LCC (h) AGC (i) PLR (j) PA (k) PA with reduced high-pass contribution

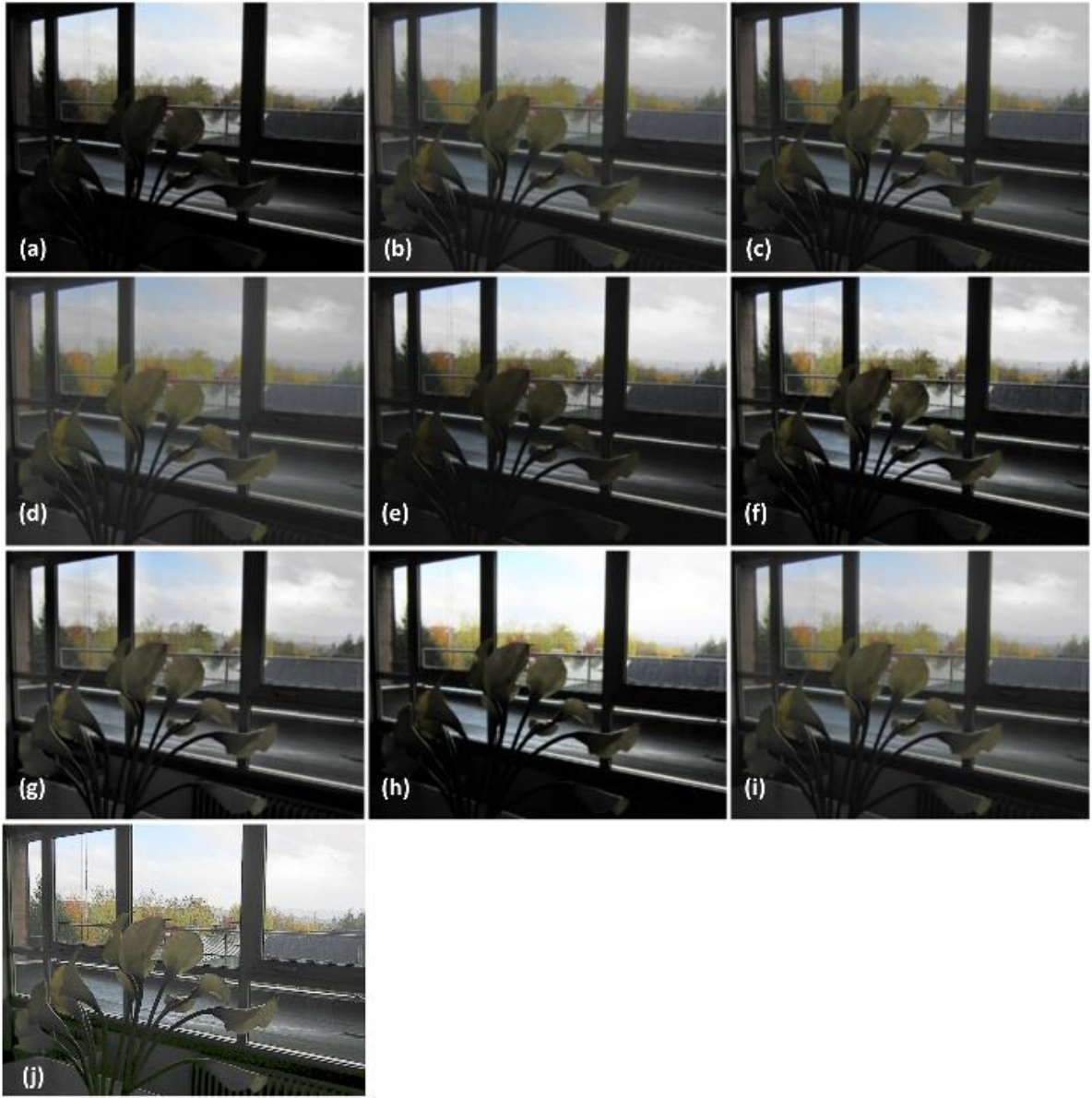

**Fig. 8** Amended figure from [8] of Iris image showing (a) original image (b) to (d) SLIP using various values (e) LCS (f) CLAHE (g) LCC (h) AGC (i) PLR (j) PA

**Table 4** Comparison of PA with algorithms from [8] for (a) *Iris* and (b) *Ferrari* images

(a)

|  | Original | SLIP $\gamma = 0.8\gamma 0$ | SLIP $\gamma = \gamma 0$ | SLIP $\gamma = 1.2\gamma 0$ | LCS | CLAHE | LCC | AGC | PLR | PA |
|---|---|---|---|---|---|---|---|---|---|---|
| σ | 0.3511 | 0.2159 | 0.2452 | 0.2693 | 0.3272 | 0.3337 | 0.3516 | **0.3885** | 0.2774 | 0.3125 |
| μ | 0.2864 | 0.3558 | 0.3384 | 0.3249 | 0.3133 | 0.3204 | 0.3049 | 0.3672 | 0.3578 | **0.4459** |
| entropy | 6.6263 | 6.534 | 6.6161 | 6.6718 | 6.4636 | 7.0151 | 7.2611 | 6.5198 | 6.6713 | **7.2988** |

(b)

|         | Original | SLIP $\gamma = 0.8\gamma_0$ | SLIP $\gamma = \gamma_0$ | SLIP $\gamma = 1.2\gamma_0$ | LCS | CLAHE | LCC | AGC | PLR | PA |
|---------|----------|------------|------------|------------|--------|--------|--------|--------|--------|--------|
| σ       | 0.2964   | 0.2376     | 0.2588     | 0.2113     | 0.3233 | 0.2907 | 0.2241 | **0.4099** | 0.3980 | 0.3023 |
| μ       | 0.2975   | 0.3238     | 0.3126     | 0.3396     | 0.3564 | 0.3341 | 0.3595 | 0.4416 | 0.3940 | **0.4610** |
| entropy | 6.6430   | 6.5893     | 6.6495     | 6.5291     | 6.6551 | **7.1288** | 6.9007 | 6.7063 | 6.7826 | 7.0031 |

### 4.3 Runtime comparison

We present the plots of the various algorithms for the 195 images used to assess the time-complexity of the algorithms and their relation to the performance of PA in Fig. 9. Based on results, PA is relatively fast in most cases, where optimization goals are quickly obtained, while in some other cases, it takes longer. Moreover, the Retinex methods exhibit the longest runtimes and this increases drastically for much larger images as seen. Thus, PA outperforms several of the closed-form approaches in terms of results and time complexity if run for a single iteration or in its adaptive mode. Additionally, it can be modified to mimic the results of other algorithms. The addition of a previously developed colour enhancement function can improve its colour rendition capabilities for certain images with faded colours. its automated, adaptive, metric optimization-based approach is advantage in some cases and its disadvantage in others. Furthermore, we have presented a successful PDE-based re-interpretation of illumination correction, dynamic range compression and contrast enhancement, easily achieved using spatial operators. Future work would involve implementing the PDE-based approach on an FPGA fabric and on a mobile embedded device platform.

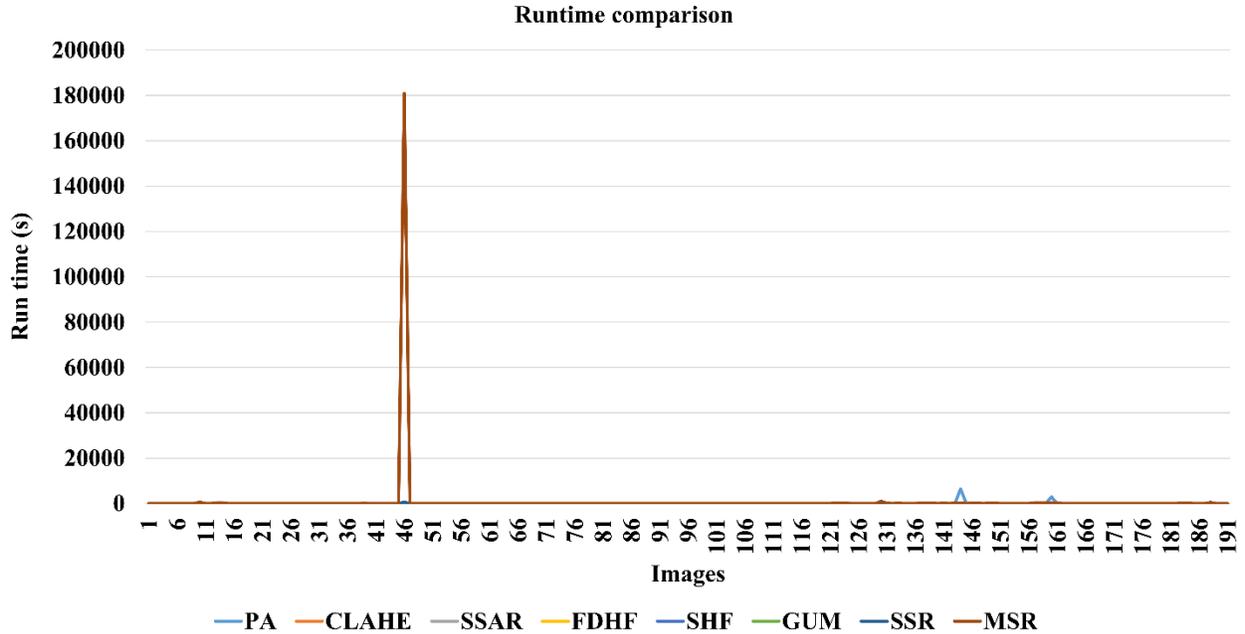

(a)

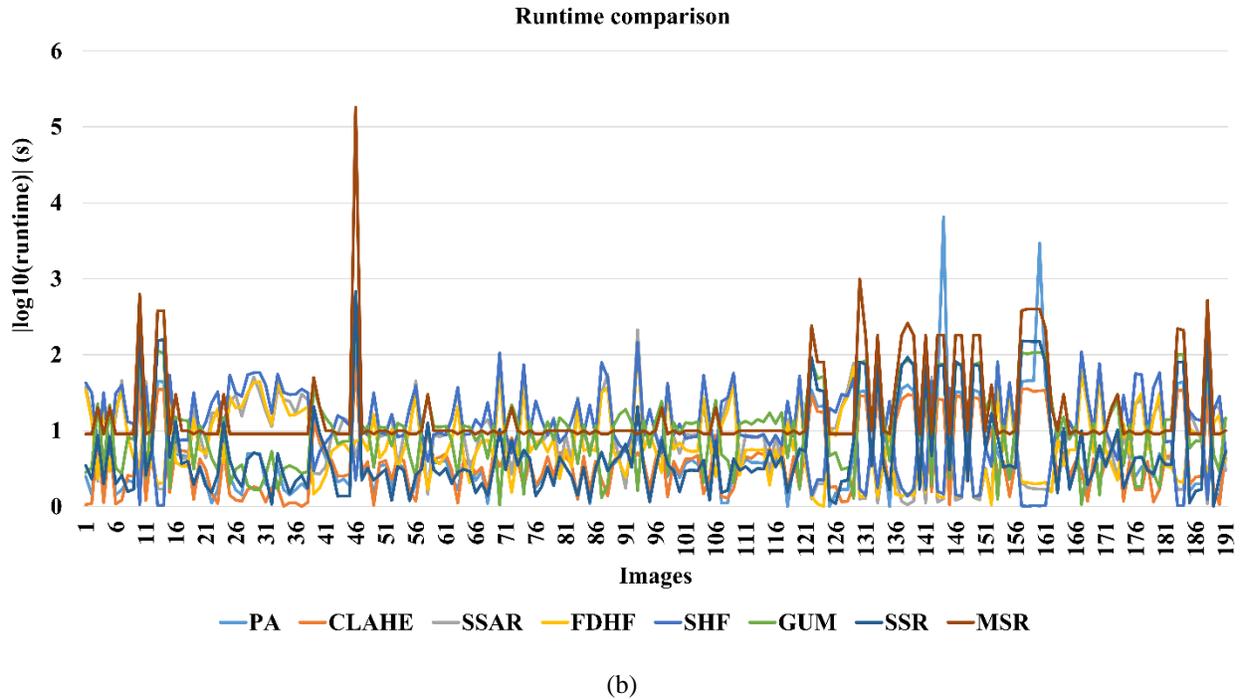

(b)

Fig. 9 Runtime comparison (a) linear (b) log domain

## 5. Conclusion

This report has presented the results of a PDE-based log-agnostic algorithm using spatial filter operators translated as opposing PDE-based flows [51]. The algorithm is of lower complexity when compared with algorithms such as CLAHE, GUM, SSR, and MSRCR. The proposed approach has been compared with numerous images and algorithms from the literature and show comparable performance with algorithms utilizing the logarithmic function and more complex operations. However, more work is needed to incorporate a strong local contrast enhancement operator that neither adds to the complexity nor over-enhances local regions, resulting in halos. Additionally, issues such as reduction of run-time and computational complexity of the algorithms in addition to improved image results will be the main focus in future work.